\documentclass{article}

\usepackage[main, final]{neurips_2026}

\usepackage[utf8]{inputenc} 
\usepackage[T1]{fontenc}    
\usepackage[colorlinks = true,
            linkcolor = black,
            citecolor = black,
            urlcolor  = olive]{hyperref}
\usepackage{url}            
\usepackage{booktabs}       
\usepackage{amsfonts}       
\usepackage{nicefrac}       
\usepackage{microtype}      
\usepackage{xcolor}         
\usepackage{booktabs}       
\usepackage{multirow}       
\usepackage{makecell}       
\usepackage{graphicx}
\usepackage{enumitem}
\usepackage{amsmath}
\usepackage{booktabs} 
\usepackage{tabularx}  
\usepackage{array}     
\usepackage{algorithm}
\usepackage{algpseudocode}
\usepackage{graphicx}
\usepackage{bm}
\usepackage{subcaption}
\RequirePackage{xspace}
\usepackage{wrapfig}
\usepackage[table]{xcolor}
\makeatletter
\DeclareRobustCommand\onedot{\futurelet\@let@token\@onedot}
\def\@onedot{\ifx\@let@token.\else.\null\fi\xspace}
\usepackage{xcolor}

\def\eg{\emph{e.g}\onedot} 

\def\ie{\emph{i.e}\onedot}

\makeatother
\newcommand{\datasetname}{Read-Parse-Describe}
\newcommand{\adaptername}{SARA\xspace}
\newcommand{\na}{\texttt{-{}-}}
\usepackage{graphicx}
\usepackage[export]{adjustbox}

\title{Read, Parse, Describe: Unified Document Parsing with Visual Element Description Generation}

\author{%
  \textbf{%
    Yuheng Chen\textsuperscript{1},~
    Yufan Chen\textsuperscript{1,}\thanks{Corresponding author:
      Yufan Chen (\texttt{yufan.chen@kit.edu}).},~
    Zhuojun Cai\textsuperscript{1},~
    Ruiping Liu\textsuperscript{1},~
    Junwei Zheng\textsuperscript{1},~
    Jiale Wei\textsuperscript{1}}\\
  \textbf{%
    Di Wen\textsuperscript{1},~
    Weijia Fan\textsuperscript{1,3},~
    Kunyu Peng\textsuperscript{1},~
    Jiaming Zhang\textsuperscript{2},~
    Rainer Stiefelhagen\textsuperscript{1}}\\
  \textsuperscript{1}Karlsruhe Institute of Technology
  \quad
  \textsuperscript{2}Hunan University
  \quad
  \textsuperscript{3}University of Alberta\\
  \texttt{chenc6417@gmail.com, yufan.chen@kit.edu}
}

\begin{document}

\renewcommand\twocolumn[1][]{#1}%
\maketitle
\begin{center}
    \centering
    \captionsetup{type=figure}
    \centering
    \begin{subfigure}[c]{0.6\textwidth}
        \centering
        \adjustimage{width=\textwidth}{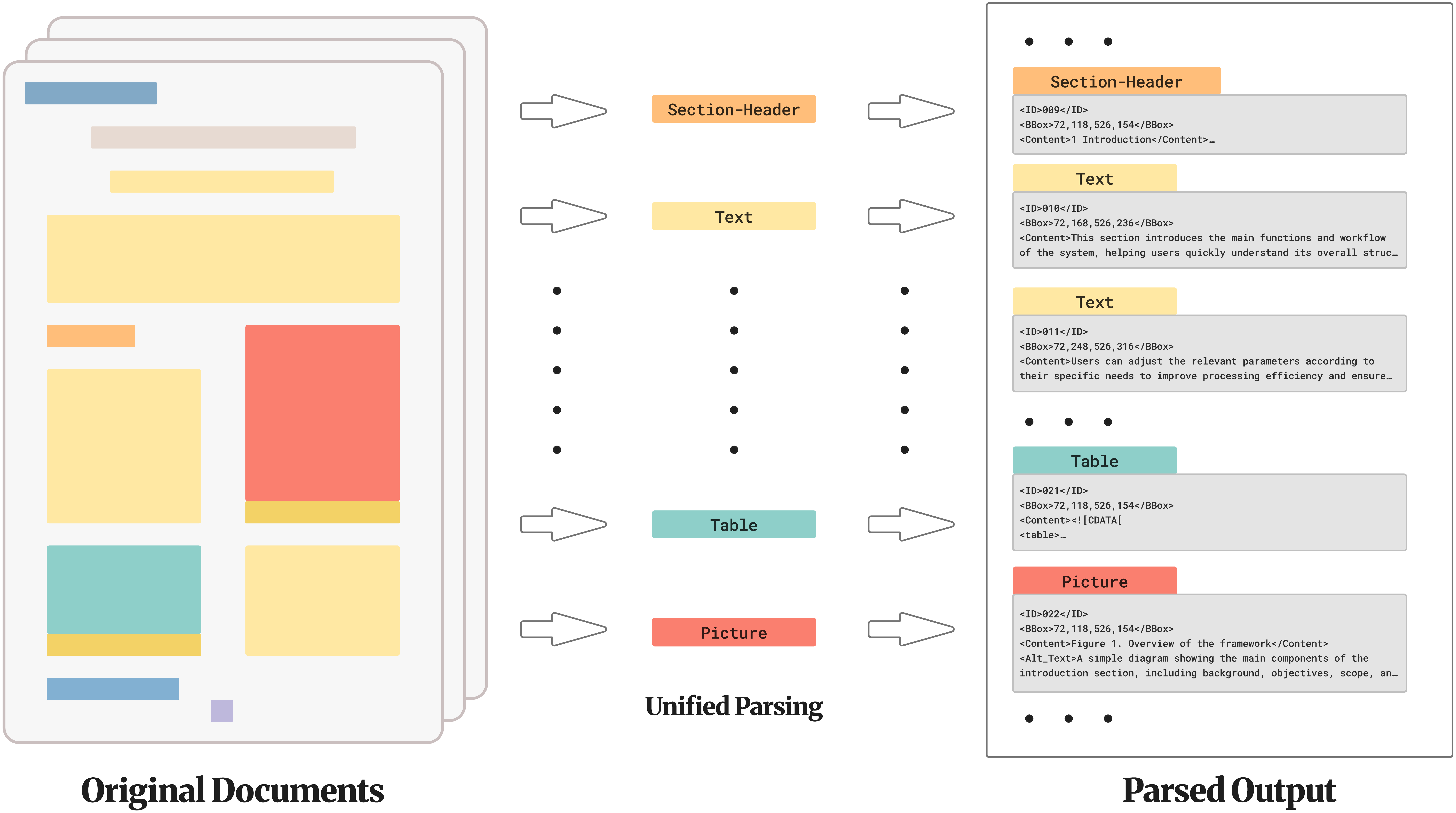}
        \caption{Unified document parsing with different document elements.} \label{fig1-a}
    \end{subfigure}
    \begin{subfigure}[c]{0.392\textwidth}
        \centering
        \includegraphics[width=1\textwidth]{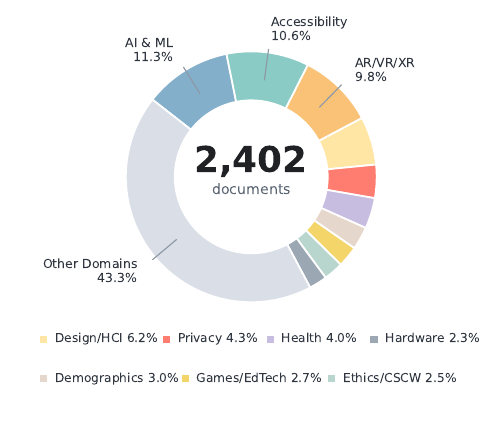}
        \caption{Domain distribution.} \label{fig1-b}
    \end{subfigure}
    \setcounter{figure}{0} 
    \captionof{figure}{\textbf{ \datasetname~Dataset Overview}. Fig.~\ref{fig1-a} shows that the unified parsing contract converts documents into structured outputs with localized elements, reading order, textual content, and visual descriptions. Fig.~\ref{fig1-b} shows the domain distribution of the 2,402 source documents. }
    \label{fig1:banner}
\end{center}%

\begin{abstract}
Document understanding systems are typically evaluated as separate components via layout analysis, optical text recognition, reading-order prediction, and image captioning. This separation leaves scientific figures, captions, and visual semantics weakly connected, and fails to measure whether a model can produce a single page-level parse that is both structurally faithful and semantically accessible. We introduce Read-Parse-Describe, a unified dataset for document parsing with visual element description generation. Given a page image, models are required to output an ordered list of layout elements containing bounding boxes, semantic labels, reading-order indices, recognized textual content, and alternative text descriptions for visual elements. The benchmark is built from 2,402 documents, comprising 18,163 continuous pages, 125,218 layout elements, and 11,338 visual instances. Besides, we further propose SARA, a plug-and-play space-awareness enhancer that reuses multi-layer visual features and injects grid-aware spatial guidance into vision-language tokens for layout-aware generation. Experiments across end-to-end VLMs and pipeline parsers show that existing systems still struggle to jointly recover reading order and visual semantics, while our enhanced model achieves strong joint performance. These results highlight the need to move beyond isolated document parsing modules toward unified models that jointly read, parse, and describe complex documents. The new dataset and code will be made publicly available at \href{https://github.com/yuheng98chen/RPD-Unidoc}{RPD Project}.
\end{abstract}

\section{Introduction}
\label{sec:introduction}

Scientific documents are not just sequences of recognized words. A single page may contain two-column prose, section headers, equations, tables, figures, captions, footnotes, author metadata, and visual evidence whose meaning is not recoverable from optical character recognition (OCR) alone. Document AI has therefore progressed through layout analysis~\cite{xu2020layoutlm,xu2021layoutlmv2,huang2022layoutlmv3,chen2024rodla}, table recognition~\cite{zhong2020imagebased}, document VQA~\cite{kim2022donut}, OCR-free parsing~\cite{wei2024gotocr,wei2025deepseekocr}, and multimodal document models~\cite{ye2023docowl}. In accessibility and machine-readable paper ingestion, the useful output is not an isolated layout map or a free-form page summary, but a structured page representation that can be read, reflowed, indexed, and audited.

Existing benchmarks usually evaluate pieces of this representation in isolation. Layout datasets measure region localization~\cite{zhong2019publaynet,li2020docbank,pfitzmann2022doclaynet}, OCR and table benchmarks~\cite{smock2022pubtables} measure textual structure, reading-order datasets~\cite{wang2021layoutreader} focus on sequencing, captioning, and alt-text datasets~\cite{sharma2018conceptualcaptions,srivatsan2024alttextcontext} describe isolated or context-conditioned images. These tasks are necessary, but they do not test whether a system can produce a single consistent parse in which boxes, labels, reading order, text, tables, captions, and visual descriptions agree with each other. This gap is particularly visible in scientific PDFs, where captions frequently refer to nearby figures or tables, equations interrupt prose, and visual elements carry technical information that should be exposed to downstream readers.

We introduce \textbf{\datasetname{}}, a unified task and dataset for unified document parsing with visual element description generation as shown in Fig.~\ref{fig1:banner}. Given a page image, a model must output an ordered list of elements, which includes a bounding box, semantic category, reading-order index, recognized text content, an explicit caption link when applicable, and an alternative-text description for visual elements. The key novelty is not any single field, but enforcing their agreement in one auditable page parse. The unified output contract can be implemented by either end-to-end models or modular systems. The benchmark contains 2,402 open-access scientific documents, 18,163 continuous pages, 125,218 layout elements, and 11,338 visual instances.

Modern VLMs and document-specialized models~\cite{dosovitskiy2020image,li2023blip2,dai2023instructblip,liu2023llava,bai2023qwenvl,lv2023kosmos25,bai2025qwen25vl,gemmateam2025gemma3} increasingly use frozen or partially frozen visual encoders. When such models merge visual tokens before generation, local spatial evidence can be weakened exactly where element extents, reading order, and caption-object alignment are required. We therefore introduce \textbf{\adaptername}, a lightweight Spatial Alignment Residual Adapter for end-to-end VLM parsing. \adaptername freezes the base VLM, reuses intermediate visual encoder layers, projects their concatenated features into the token space, aligns with crop-aware spatial grids, and injects into the native merged tokens. The task makes spatial-alignment failures measurable, while \adaptername tests a direct architectural response without changing labels or evaluation. To summarize, our contributions are as follows:
\begin{itemize}
    \item We introduce \datasetname{}, a novel task and dataset for unified document parsing with alternative-text generation into a shared structured schema.
    \item We propose \adaptername{}, a Spatial Alignment Residual Adapter that injects crop-aware spatial cues into VLM tokens to improve layout-sensitive document understanding.
    \item We establish a unified evaluation protocol covering both document parsing and alternative-text generation, enabling systematic assessment of parsing fidelity and generation quality.
    \item We conduct extensive experiments and ablation studies, showing that crop-aware spatial alignment consistently benefits unified document parsing and alternative-text generation.
\end{itemize}

\section{Related Work}
\label{sec:related_work}

\subsection{Document Layout and Parsing Datasets}
Document-layout benchmarks have made document understanding measurable by isolating the output. Early resources, such as RVL-CDIP~\cite{harley2015rvlcdip}, treat documents as classes, whereas FUNSD~\cite{jaume2019funsd} focuses on form. CORD~\cite{park2019cord} and SROIE~\cite{huang2021sroie} further show that domain-specific schemas enhance extraction but are specialized. Similar to the table, SciTSR~\cite{chi2019scitsr} targets scientific table structure, and PubTables-1M~\cite{smock2022pubtables} scales table recognition, but neither requires the table, caption, reading position, and visual description to agree in one element record.  PubLayNet~\cite{zhong2019publaynet} made large-scale layout detection practical, and DocBank~\cite{li2020docbank} expanded token-level layout supervision. DocLayNet~\cite{pfitzmann2022doclaynet} improved further with richer human annotations, while M6Doc~\cite{cheng2023m6doc} expanded layout analysis across heterogeneous documents. For robustness, RoDLA~\cite{chen2024rodla} stresses layout robustness, whereas ReadingBank~\cite{wang2021layoutreader} isolates reading-order prediction. Broader evaluations such as DocVQA~\cite{mathew2021docvqa} and DUDE~\cite{vanlandeghem2023dude} test document understanding through answers rather than page parses. OmniDocBench~\cite{ouyang2025omnidocbench} moves closer to document parsing, while MMLongBench-Doc~\cite{ma2024mmlongbenchdoc} stresses long-context document understanding.

\subsection{Document VLMs and Pipeline Systems}
Structure-aware representations are often precise, while generative systems are flexible but low-performing. LayoutLM~\cite{xu2020layoutlm} and LayoutLMv2~\cite{xu2021layoutlmv2} made layout-aware pretraining central for document understanding. LayoutLMv3~\cite{huang2022layoutlmv3} further unifies text and image modeling through task-specific heads rather than a complete generator. DocFormer~\cite{appalaraju2021docformer} and VGT~\cite{da2023vgt} emphasize structure-aware encoding, reinforcing the explicit geometry. OCR-free generation such as Donut~\cite{kim2022donut} can emit structured text directly from an image, while Pix2Struct~\cite{lee2023pix2struct} supports screenshot-style pretraining. Nougat~\cite{blecher2023nougat} specializes this generative view for academic documents, and GOT-OCR~\cite{wei2024gotocr} extends OCR-style generation toward broader text recognition. Recent document parsers push toward full-page extraction, \eg, DeepSeek-OCR~\cite{wei2025deepseekocr} uses visual compression for OCR, while Dolphin-v2~\cite{feng2026dolphinv2} illustrates universal document parsing. MonkeyOCR~\cite{li2025monkeyocr} emphasizes structure-recognition-relation decomposition, and PaddleOCR-VL~\cite{cui2025paddleocrvl} targets compact multilingual document parsing. MinerU~\cite{wang2024mineru} takes a more inspectable pipeline route. General VLMs provide another axis of flexibility, \eg, BLIP-2~\cite{li2023blip2} and InstructBLIP~\cite{dai2023instructblip} align visual features with language models, while LLaVA~\cite{liu2023llava} and Qwen2.5-VL~\cite{bai2025qwen25vl} show strong instruction-following multimodal generation. Agentic parsing is also studied in ParseFixer~\cite{yu2026parsefixer}, which applies selective correction to backbone Markdown outputs, and ParseBench~\cite{zhang2026parsebench}, which evaluates agent-relevant parsing capabilities on enterprise documents.

\subsection{Visual Descriptions and Accessibility}
Accessibility work shows why visual description cannot be reduced to generic captioning. Conceptual Captions~\cite{sharma2018conceptualcaptions} demonstrates the scale available in web alt text, but weakly tied to accessibility intent. VizWiz~\cite{gurari2018vizwiz} and VizWiz-Captions~\cite{gurari2020captioning} foreground blind-user scenarios, where the useful answer depends on context. TextCaps~\cite{sidorov2020textcaps} shows that descriptions may depend on visible text, and OCR-VQA~\cite{mishra2019ocrvqa} makes text reading part of visual question answering. Screen2Words~\cite{wang2021screen2words} shifts the target to UI screens, while Alt-Text with Context~\cite{srivatsan2024alttextcontext} conditions descriptions on surrounding social text. Furthermore, ChartQA~\cite{masry2022chartqa} requires visual and logical reasoning, and DePlot~\cite{liu2023deplot} recovers chart data before reasoning. Chart-to-Text~\cite{kantharaj2022charttotext} directly studies chart description, while PlotQA~\cite{methani2020plotqa} emphasizes quantitative plot understanding. A useful description must be tied to the correct object, caption, page position, and surrounding reading flow. \datasetname{} therefore evaluates alternative text as a field in the parse rather than as an isolated captioning output.

\section{Read-Parse-Describe Dataset}
\label{sec:benchmark}

\subsection{Task Definition}
\label{subsec:task_definition}
We introduce Read-Parse-Describe, a dataset for document parsing with grounded visual description. This task is motivated by two limitations of existing document understanding benchmarks, as shown in Tab.~\ref{tab:dataset_comparison}. First, many layout datasets primarily focus on the physical localization of page regions but do not require models to recover a complete logic-aware document representation. This limitation is especially pronounced in scientific papers, where simple top-to-bottom ordering rules are frequently broken. Second, visual elements are often treated as layout categories rather than semantic objects. Figures, charts, diagrams, and screenshots may be localized, but the information they convey is rarely required to be presented in a grounded, accessible form.

\begin{table}[t]
\centering
\scriptsize
\setlength{\tabcolsep}{2.0pt}
\renewcommand{\arraystretch}{0.92}
\caption{
Comparison of dataset target contracts. A checkmark indicates that the benchmark explicitly requires the corresponding field as part of its evaluation target. \textbf{Caption Links} refer to explicit associations between visual objects and their captions. \textbf{Structured Output} denotes a single page-level prediction that jointly contains layout, order, textual content, and caption links.
}
\label{tab:dataset_comparison}
\resizebox{\textwidth}{!}{%
\begin{tabular}{lcccccccc}
\toprule
\textbf{Dataset} & \textbf{Docs} & \makecell{\textbf{Layout}} & \makecell{\textbf{Reading}\\\textbf{Order}} & \makecell{\textbf{Textual}\\\textbf{Elements}} & \makecell{\textbf{Visual}\\\textbf{Elements}} & \makecell{\textbf{Caption}\\\textbf{Links}} & \makecell{\textbf{Alternative}\\\textbf{Text}} & \makecell{\textbf{Structured}\\\textbf{Output}} \\
\midrule
PubLayNet~\citep{zhong2019publaynet} & PDF & \checkmark & -- & -- & \checkmark & -- & -- & -- \\
DocLayNet~\citep{pfitzmann2022doclaynet} & Mixed & \checkmark & -- & -- & \checkmark & -- & -- & -- \\
DocBank~\citep{li2020docbank} & PDF & \checkmark & -- & -- & \checkmark & -- & -- & -- \\
M6Doc~\citep{cheng2023m6doc} & Mixed & \checkmark & -- & -- & \checkmark & -- & -- & -- \\
D4LA~\citep{da2023vgt} & Mixed & \checkmark & -- & -- & \checkmark & -- & -- & -- \\
SciTSR~\citep{chi2019scitsr} & Images & -- & -- & \checkmark & -- & -- & -- & -- \\
ReadingBank~\citep{wang2021layoutreader} & Mixed & \checkmark & \checkmark & \checkmark & -- & -- & -- & -- \\
DocVQA~\citep{mathew2021docvqa} & Images & -- & -- & \checkmark & \checkmark & -- & -- & -- \\
ChartQA~\citep{masry2022chartqa} & Images & -- & -- & -- & \checkmark & -- & -- & -- \\
InfographicVQA~\citep{mathew2022infographicvqa} & Images & -- & -- & -- & \checkmark & -- & -- & -- \\
DUDE~\citep{vanlandeghem2023dude} & Mixed & -- & -- & \checkmark & \checkmark & -- & -- & -- \\
OmniDocBench~\citep{ouyang2025omnidocbench} & Mixed & \checkmark & \checkmark & \checkmark & \checkmark & \checkmark & -- & \checkmark \\
\midrule
\textbf{Read-Parse-Describe (Ours)} & \textbf{PDF} & \textbf{\checkmark} & \textbf{\checkmark} & \textbf{\checkmark} & \textbf{\checkmark} & \textbf{\checkmark} & \textbf{\checkmark} & \textbf{\checkmark} \\
\bottomrule
\end{tabular}%
}
\end{table}

\subsection{Output Schema}
\label{subsec:output_schema}
Read-Parse-Describe therefore evaluates document understanding as a unified multi-modal parsing problem. The target output is an ordered list of document elements. Each element contains its bounding box, semantic category, reading-order index, recognized content when applicable, caption linkage, and visual description when applicable:
\begin{equation}
Y = [e_1, e_2, \ldots, e_n],
\end{equation}
where each element is represented as
\begin{equation}
e_i = \{\texttt{id}, \texttt{bbox}, \texttt{category}, \texttt{order}, \texttt{text}, \texttt{html}, \texttt{caption\_id}, \texttt{alt\_text}\},
\end{equation}
where \texttt{id} is the element identifier, \texttt{bbox} is a page-coordinate box $[x_1,y_1,x_2,y_2]$ in the rendered image coordinate system. \texttt{category} is the semantic layout label, and \texttt{order} is the canonical human reading-order index. The submitted list is expected to be sorted by \texttt{order}, \ie, if list position and \texttt{order} disagree, evaluation first normalizes by the \texttt{order} field and leaves duplicate or missing indices to be penalized by matching and sequence metrics. \texttt{text} stores recognized text for textual regions, \texttt{html} stores table structure, \texttt{caption\_id} links a visual object to caption, and \texttt{alt\_text} stores a factual visual description. Non-applicable fields are set to \texttt{null}.

\paragraph{Terminology.}
An \emph{ordered schema} is the full element list $Y$ after sorting by the reading-order index. 
The central design choice is to make fields agree inside the same element object, not to score familiar subtasks in isolation. The contract has four coupled decisions:
{
\begin{center}
\vspace{-0.35em}
\scriptsize
\setlength{\tabcolsep}{3.0pt}
\renewcommand{\arraystretch}{1.05}
\begin{tabularx}{\textwidth}{@{}p{2.55cm}p{4.35cm}X@{}}
\toprule
\textbf{Coupling} & \textbf{Design reason} & \textbf{Failure exposed} \\
\midrule
Element identity & Bind \texttt{id}, \texttt{category}, textual content, caption links, and descriptions to the same object. & Free-floating OCR, HTML, captions, or descriptions cannot be credited without the element they claim to describe. \\
Geometry as evidence & Treat \texttt{bbox} as the audit handle for text, tables, and visual descriptions. & A plausible description or table string attached to the wrong crop becomes measurable rather than silently accepted. \\
Canonical reading flow & Use \texttt{order} as human reading order rather than visual-nearest-neighbor order. & Two-column text, spanning figures, and separated captions expose sequence errors even when OCR is correct. \\
Caption semantics & Require \texttt{caption\_id} and \texttt{alt\_text} for visual instances when applicable. & A model must bind the object, caption, and description instead of generating generic image captions. \\
\bottomrule
\end{tabularx}
\vspace{-0.55em}
\end{center}
}

The couplings above make the task a test of \emph{spatially grounded binding}. Region proposals decide the visual evidence available to OCR, table serialization, caption linkage, and alt-text generation. Reading-order decisions decide where captions and objects sit in the page narrative; and visual descriptions are auditable only when their evidence is localized.

\subsection{Dataset Creation Pipeline}
\label{subsec:dataset_creation}

We collect open-access PDF papers from the ACM Digital Library and retain document metadata. Since the main contributions, core figures, methods, and experimental results of scientific papers are usually concentrated in the main body, we extract the first eight pages of each paper and keep shorter papers in their entirety. This page-level curation strategy preserves structurally and semantically informative regions of papers while reducing the dominance of long reference sections and appendices. The dataset presents a more balanced distribution across diverse scientific disciplines without diluting evaluation on repetitive textual lists. 
Additional detailed dataset statistics would be included in the technical appendix.

The visual elements include tables, diagrams, screenshots, photographs, charts, and long-tail illustrations. This diversity matters because visual-description generation is not the same problem for each object type, \eg, a UI screenshot requires interface semantics. The layout labels are similarly heterogeneous, covering structural headers, scholarly objects, captions, references, lists, and paragraphs. We keep these categories explicit to avoid hiding errors that make scientific PDFs hard to parse. We split at the paper level with a 90:5:5 train/validation/test ratio after deduplication. Exact duplicate PDFs are removed using SHA-256 hashes, DOI metadata is used when available, normalized titles are matched otherwise, and near-duplicate title/abstract pairs above the similarity threshold are manually inspected. Paper-level splitting is important because adjacent pages share templates, section styles, figure numbering, author metadata, and sometimes repeated visual objects. A page-level split would overstate generalization for reading order and caption linkage.

\subsection{Annotation Pipeline}
\label{subsec:annotation_pipeline}

Annotations are produced with a bootstrap-and-correct pipeline designed around cross-field consistency. A document parser first initializes layout boxes and labels. The generated boxes, semantic categories, caption links, and reading-order indices are then manually inspected and corrected together, rather than as independent layers. Corrections target common scientific-PDF failure modes: two-column reading order, spanning figures, captions separated from visual objects, footnotes, equations, author metadata, and fine-grained heading levels. This joint correction step is important: moving a box can change its text crop, its caption candidate, and its reading-order position.

Visual instances then receive alternative-text annotations through a triage loop: generate an initial description, score it against the accessibility rubric, refine low-score cases by human review, and audit the revised labels. Gemini 3 Pro~\cite{googledeepmind2025gemini3pro} provides the initial descriptions and Gemini 3 Flash~\cite{googledeepmind2025gemini3flash} scores factual accuracy, completeness, and prompt adherence on a 1--5 scale. The 1,408 descriptions scoring below 4 are routed to human refinement, then re-scored:

\begin{wraptable}{l}{0.5\linewidth}
\vspace{-1.2em}
\scriptsize
\setlength{\tabcolsep}{3.2pt}
\renewcommand{\arraystretch}{1.02}
\begin{tabular}{@{}lcc@{\qquad}lcc@{}}
\toprule
\textbf{Score} & \textbf{Generated} & \textbf{Refined} &
\textbf{Score} & \textbf{Generated} & \textbf{Refined} \\
\midrule
1 & 0 & 0 & 4 & 2,419 & 226 \\
2 & 146 & 0 & 5 & 7,511 & 11,112 \\
3 & 1,262 & 0 & \textbf{Score-5 rate} & \textbf{66.2\%} & \textbf{98.1\%} \\
\bottomrule
\end{tabular}
\vspace{-0.55em}
\end{wraptable}

After refinement, the score-5 rate increases from 66.2\% to 98.1\%. A manual audit of 200 revised samples with previous scores below 5 gives 194/200 exact agreement between human and LLM scores. The technical appendix includes the prompt, scoring rubric, full score distribution, and audit matrix for this annotation stage.

The alt-text process is designed for benchmark consistency, not as an automatic accessibility deployment pipeline. The prompt asks for concise factual descriptions and discourages speculation beyond visible evidence. The scoring step checks whether the description is faithful, complete enough for the object type, and aligned with the prompt; human refinement is concentrated on low-score cases because these are most likely to contain missing relationships, hallucinated labels, or vague descriptions. We expose this procedure because visual-description labels are harder to audit than boxes or OCR strings, and the reader should be able to separate measured model behavior from annotation assumptions.

\section{\adaptername: Spatial Alignment Residual Adapter}
\label{sec:method}

\datasetname{} turns parsing into a binding problem, \ie, each generated element must keep the attribute attached to the same page evidence. This creates a different pressure from ordinary VLM generation. A visual concatenation is useful due to the compression of image tokens into a sequence the decoder can consume, but that compression can blur the local spatial cues needed to decide where an element starts, how two-column regions should be ordered, and which caption belongs to which object. The design goal of \adaptername{} is therefore not to replace the base VLM, but to restore crop-level spatial evidence at the point where the decoder receives merged visual tokens. The key design choice is to preserve correspondence between each crop's pre-merger grid and its native post-merger tokens, enabling aligned feature re-injection without ground-truth input boxes.

\section{Method}
\begin{figure}
   \centering
    \includegraphics[width=0.98\textwidth]{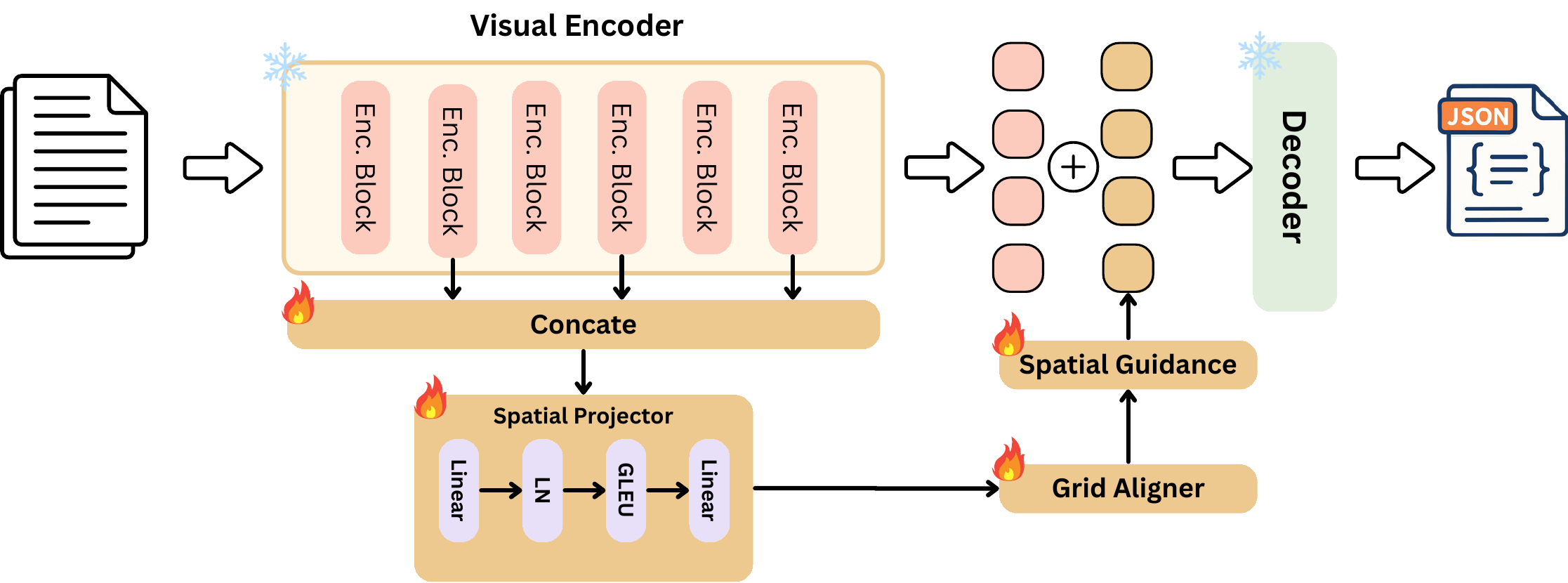}
     \caption{Document parsing model with SARA adapter, which is a plug-and-play spatial enhancer that leverages multi-layer visual features. Features from encoder layers are concatenated, followed by a grid-aware multi-crop alignment mechanism that preserves spatial structure. The aligned features serve as spatial guidance and are injected into the original tokens via a gated residual connection.}
     \label{fig:our model}
\end{figure}

\adaptername{} follow threefold design as shown in the Fig.~\ref{fig:our model}:
(1) Reuse intermediate visual states rather than only the final merged tokens, because earlier layers retain finer layout and boundary information.
(2) Align the reused states to the same crop grid as the native visual merger, so the residual signal has the same token-level support as the decoder's original visual stream.
(3) Inject the aligned guidance through a learnable residual gate, preserving the pretrained visual-language pathway while letting training decide how much spatial correction is useful.

\paragraph{Multi-layer spatial features.}
Let $\mathcal{S}=\{2,5,8,11\}$ be the selected visual encoder layers. During the forward pass, hooks collect hidden states
\begin{equation}
\bm X^{(l)} \in \mathbb{R}^{L_v \times d_v}, \quad l \in \mathcal{S},
\end{equation}
where $L_v$ is the pre-merger visual-token length and $d_v$ is the visual feature dimension. We concatenate these states along the channel dimension:
\begin{equation}
\bm X_{\mathrm{cat}} = \operatorname{Concat}\left(\bm X^{(2)},\bm X^{(5)},\bm X^{(8)},\bm X^{(11)}\right).
\end{equation}

The concatenated feature is mapped to the merger dimension $d_m$ by a small projector,
\begin{equation}
\bm G = \bm W_2\,\sigma\left(\operatorname{LN}(\bm W_1 \bm X_{\mathrm{cat}})\right),
\end{equation}
where $\operatorname{LN}$ is LayerNorm and $\sigma$ is GELU.

\paragraph{Crop-aware alignment.}
The visual merger reduces spatial resolution, so the guidance $\bm G$ must be aligned with the native merged tokens. For each crop with grid metadata $(t_i,h_i,w_i)$, we slice the corresponding segment from $\bm G$, reshape it into its two-dimensional grid, and average-pool to the merged-grid resolution:
\begin{equation}
\bm G_i'=\operatorname{Pool}_{2D}\left(\operatorname{Reshape}(\bm G_i), (h_i/2,w_i/2)\right).
\end{equation}
All crop-level features are flattened and concatenated into $\bm G' \in \mathbb{R}^{L_m \times d_m}$, matching the length and dimension of the original merged visual tokens $T$.

\paragraph{Residual injection.}
The enhanced visual tokens are
\begin{equation}
\hat{\bm  T}=\bm T+\lambda \bm G',
\end{equation}
where $\lambda$ is a learnable scalar gate initialized to 1.0. The language decoder then generates the same ordered schema used by all baselines. \adaptername changes only the visual token pathway; it does not alter the benchmark labels, target schema, or evaluation metrics.

\paragraph{Shape summary and scope.}
The data flow is $\bm X^{(l)}\!\in\!\mathbb{R}^{L_v\times d_v}$, $\bm  X_{\mathrm{cat}}\!\in\!\mathbb{R}^{L_v\times |\mathcal{S}|d_v}$, $\bm G\!\in\!\mathbb{R}^{L_v\times d_m}$, and $\bm G',\bm T,\hat{\bm T}\!\in\!\mathbb{R}^{L_m\times d_m}$ after crop-aware pooling. \adaptername is not intended to replace document-specific supervision or table-aware decoding. The residual form keeps the original visual-token stream intact and lets training learn the aligned guidance.

\section{Experiments}
\label{sec:experiment}

\subsection{Settings and Metrics}
\label{subsec:settings_metrics}

We evaluate three categories of systems: general-purpose VLMs, specialized document models, and pipeline parsers. General-purpose VLMs include Qwen3.5~\cite{qwen}, Gemma4~\cite{gemma4}, ChatGPT-5.4~\cite{openai2026gpt54}, ChatGPT-5.6~\cite{openai2026gpt56}, Gemini 3.6 Flash~\cite{google2026gemini36flash} and Gemini 3.1 Pro~\cite{google2026gemini31pro}. Specialized document models include DeepSeek-OCR~\cite{wei2025deepseekocr}, DeepSeek-OCR2~\cite{deepseekocr2}, Dolphin-v2~\cite{feng2026dolphinv2}, MonkeyOCR~\cite{li2025monkeyocr}, VGT~\cite{da2023vgt}, GOT-OCR 2.0~\cite{wei2024gotocr}, Nougat~\cite{blecher2023nougat}, and PaddleOCR-VL~\cite{cui2025paddleocrvl}. Pipeline parsers include PP-StructureV3~\cite{paddleocr3} and MinerU~\cite{wang2024mineru}.

All systems are evaluated under the same page-level output schema. Zero-shot systems use released checkpoints or official APIs without in-context examples. Full fine-tuning baselines serialize each target element list into text and optimize the standard autoregressive next-token objective on the training split. The validation split is used for model selection, and final results are reported on the held-out test split. For \adaptername{} runs, the base VLM is frozen and only the adapter pathway is trained. For generative VLMs, we use deterministic decoding whenever supported. Outputs from pipeline systems are converted into the unified schema.

\paragraph{Metrics.}
All metrics are computed after schema conversion. Minor syntax errors are repaired deterministically; if a prediction remains unparsable, the affected fields are treated as empty. F1 follows the IoU-matching convention used by PubLayNet~\cite{zhong2019publaynet}, DocLayNet~\cite{pfitzmann2022doclaynet}, and RoDLA~\cite{chen2024rodla}: a match requires the same category label and $\operatorname{IoU}\ge0.5$. mIoU is computed over matched true positives. Reading-order (RO) score uses normalized edit-distance similarity over matched element identifiers~\cite{wang2021layoutreader}. OCR accuracy is computed as normalized character-level edit-distance similarity after text normalization. For visual descriptions, we report METEOR~\cite{banerjee2005meteor} and BERTScore~\cite{zhang2020bertscore}.

\subsection{Main Results}
\label{subsec:main_results}

Table~\ref{tab:ocr_alt_eval} should be read as a contract diagnostic, not a leaderboard. The unified schema requires models to jointly recover localized elements, reading order, textual content, and visual descriptions. Pipeline parsers often retain strong OCR or localization performance, but they usually leave caption links and alt-text fields incomplete. General-purpose VLMs can produce richer schema fields, but they are less reliable in spatial grounding without adaptation. Document-specialized VLMs and \adaptername{} are complementary: the former excel at specialized parsing, while \adaptername{} improves the structural fields of a frozen general-purpose VLM and maintains competitive visual-description quality.

\renewcommand{\arraystretch}{0.9}
\begin{table*}[t]
\centering
\caption{Main test-set results under the unified \datasetname{} output schema. Rows are grouped by evaluation setting to separate \textbf{Zero-shot Models}, \textbf{Full Fine-Tuning}, which means fine-tuning the whole model, and \textbf{Adapter Tuning}, which means only fine-tuning the adapter part. \na{} indicates that the corresponding field is not produced by the evaluated system. \textbf{bold} and \underline{underlined} scores denote the best and second-best results for each metric, respectively.}
\label{tab:ocr_alt_eval}
\scriptsize
\setlength{\tabcolsep}{6.8pt}
\renewcommand{\arraystretch}{1.08}
\resizebox{0.98\textwidth}{!}{%
\begin{tabular}{@{}lccccccc@{}}
\toprule
\multirow{2}{*}[-0.6ex]{\makecell[c]{{\textbf{\quad Model}}}} &
\multirow{2}{*}[-0.6ex]{\makecell[c]{\textbf{Trainable}\\\textbf{Params}}} &
\multicolumn{2}{c}{\textbf{DLA}} &
\multirow{2}{*}{\makecell[c]{\textbf{OCR}\\\textbf{Acc.$\uparrow$}}} &
\multirow{2}{*}{\makecell[c]{\textbf{RO}\\\textbf{Score$\uparrow$}}} &
\multicolumn{2}{c}{\textbf{Alt. Text}} \\
\cmidrule(lr){3-4}
\cmidrule(lr){7-8}
&
&
\textbf{F1$\uparrow$} &
\textbf{mIoU$\uparrow$} &
&
&
\textbf{METEOR$\uparrow$} &
\makecell[c]{\textbf{BERT $\uparrow$}} \\
\midrule

\rowcolor{gray!15}\multicolumn{8}{l}{\textit{Zero-shot Models}} \\
\quad GOT-OCR 2.0 & \na{} & \na{} & \na{} & 0.8652 & 0.5330 & \na{} & \na{} \\
\quad Nougat & \na{} & \na{} & \na{} & 0.7327 & 0.4640 & \na{} & \na{} \\
\quad Dolphin-v2 & \na{} & 0.9124 & 0.9265 & 0.9307 & 0.8221 & \na{} & \na{} \\
\quad MonkeyOCR & \na{} & 0.6030 & 0.8949 & 0.9684 & 0.6254 & \na{} & \na{} \\
\quad VGT & \na{} & 0.6795 & 0.8540 & 0.6304 & 0.3833 & \na{} & \na{} \\
\quad PaddleOCR-VL & \na{} & 0.7564 & 0.9458 & 0.9043 & 0.6619 & \na{} & \na{} \\
\quad PP-StructureV3 & \na{} & 0.6375 & 0.8870 & 0.9558 & 0.5054 & \na{} & \na{} \\
\quad MinerU & \na{} & 0.7171 & 0.9151 & 0.9671 & 0.6757 & \na{} & \na{} \\
\quad ChatGPT-5.4 & \na{} & 0.6978 & 0.6454 & 0.9796 & 0.7253 & 0.3506 & 0.8751 \\
\quad ChatGPT-5.6 & \na{} & 0.7489 & 0.7781 & 0.9448 & 0.8967 & 0.3793 & \textbf{0.9004} \\
\quad Gemini 3.6 Flash & \na{} & 0.6909 & 0.8658 & 0.9544 & 0.8183 & 0.3744 & 0.8841 \\
\quad Gemini 3.1 Pro & \na{} & 0.7421 & 0.8551 & 0.9633 & 0.8337 & 0.3785 & 0.8961 \\
\quad Gemma4-E4B & \na{} & 0.5473 & 0.7674 & 0.8977 & 0.5256 & 0.1679 & 0.8455 \\
\quad DeepSeek-OCR & \na{} & 0.5987 & 0.6074 & 0.9064 & 0.5003 & 0.2562 & 0.8280 \\
\quad DeepSeek-OCR2 & \na{} & 0.5806 & 0.5963 & 0.8688 & 0.4830 & 0.2604 & 0.8332 \\

\addlinespace[0.25em]
\midrule

\rowcolor{gray!15}\multicolumn{8}{l}{\textit{Full fine-tuning}} \\
\quad Qwen3.5-0.8B & 752.39M & 0.7177 & 0.8345 & 0.9475 & 0.9605 & 0.3666 & 0.8904 \\
\quad Qwen3.5-2B & 1881.83M & 0.7336 & 0.8350 & 0.9408 & 0.9628 & 0.3741 & 0.8928 \\
\quad Gemma4-E4B & 7463.01M & 0.9358 & 0.8836 & 0.9452 & 0.9628 & 0.3047 & 0.8821 \\
\quad DeepSeek-OCR & 2937.36M & \underline{0.9644} & \underline{0.9492} & \underline{0.9679} & \underline{0.9825} & \textbf{0.3998} & \underline{0.8968} \\
\quad DeepSeek-OCR2 & 2935.88M  & 0.8783 & 0.9189 & 0.9416 & 0.9231 & 0.3229 & 0.8094 \\
\quad Dolphin-v2 & 3085.93M & \textbf{0.9826} & \textbf{0.9588} & \textbf{0.9751} & \textbf{0.9857} & 0.3396 & 0.8838 \\

\addlinespace[0.25em]
\midrule

\rowcolor{gray!15}\multicolumn{8}{l}{\textit{Adapter Tuning}} \\
\quad Gemma4-E4B + LoRA & 103.82M & 0.7687 & 0.7238 & 0.9164 & 0.9451 & 0.2871 & 0.8753 \\
\quad Gemma4-E4B + \adaptername & 158.82 M & 0.9437 & 0.9136 & 0.9474 & 0.9288 & 0.3009 & 0.8746 \\
\quad Qwen3.5-2B + LoRA & 48.37M & 0.7004 & 0.6845 & 0.9673 & 0.9304 & 0.3586 & 0.8652 \\
\quad Qwen3.5-2B + \adaptername & 83.99 M & 0.9189 & 0.8919 & 0.9636 & 0.9774 & \underline{0.3834} & 0.8946 \\

\bottomrule
\end{tabular}%
}
\vspace{-1em}
\end{table*}

Table~\ref{tab:ocr_alt_eval} reveals different strengths across the evaluated systems and reported metrics. Fine-tuned Dolphin-v2 achieves the strongest localization performance, reaching 0.9826 F1 and 0.9588 mIoU. It also obtains the highest RO score, 0.9857.  Fine-tuned DeepSeek-OCR achieves the highest METEOR, 0.3998, while GPT-5.6 achieves the highest BERTScore, 0.9004. DeepSeek-OCR also maintains a high RO score of 0.9825. Among the evaluated proprietary zero-shot models, GPT-5.6 obtains the highest F1, RO, and description scores, while Gemini 3.6 Flash obtains the highest mIoU. Including Gemini 3.1 Pro extends the comparison to the requested Pro-series model under the same schema. These results suggest that spatially precise parsing and open-ended visual description remain partially distinct capabilities.

Pipeline parsers remain competitive on some low-level fields but are less complete under the unified output schema. For example, MinerU reaches 0.9671 OCR accuracy and 0.9151 mIoU, but its RO score is 0.6757, and it does not produce visual alternative text under the evaluated schema. This does not mean that pipeline systems are ineffective. Instead, it identifies the fields that require additional modules or tighter integration when the target is a complete accessible page parse.

Compared with the Qwen3.5-2B full fine-tuning baseline, \adaptername{} substantially improves structural performance while keeping the base VLM frozen and training with 83.99M parameters. F1 increases to 0.9189, mIoU to 0.8919, OCR accuracy to 0.9636, and RO score to 0.9774. Its alt-text scores, 0.3834 METEOR and 0.8946 BERTScore, are also competitive with the strongest fine-tuned document VLMs.  On Gemma4-E4B, \adaptername{} also improves F1 and mIoU relative to full fine-tuning, but RO and both description metrics decrease slightly. Compared with LoRA-only adaptation, \adaptername{} improves F1 and mIoU by 0.2185 and 0.2074 on Qwen and 0.1750 and 0.1898 on Gemma, respectively. Other metrics show trade-offs: Qwen LoRA has higher OCR accuracy, while Gemma LoRA has higher RO and slightly higher BERTScore.

\paragraph{Interpreting the joint table.}
The main result is not that one model is uniformly best, but that single-axis strengths do not necessarily imply a coherent page parse. A model with high OCR but low RO may still fail to recover columns or reading sequences. A model with strong F1 but no alt-text may be useful as a front end, but it is incomplete as an accessible parser. Conversely, a model with strong alt-text but weak mIoU may describe objects that are poorly localized. Reporting these fields side by side makes such tradeoffs visible without imposing a deployment-specific ranking.

\subsection{Ablations and Analysis}
\label{subsec:insights}

\begin{table}[t]
\centering
\begin{minipage}[t]{0.49\textwidth}
\centering
\caption{Component ablation with Qwen3.5-2B.}
\label{tab:spatial_ablation}
\scriptsize
\setlength{\tabcolsep}{2.4pt}
\renewcommand{\arraystretch}{0.92}
\resizebox{\linewidth}{!}{%
\begin{tabular}{@{}lccccc@{}}
\toprule
\textbf{Configuration} &
\makecell[c]{\textbf{Trainable}\\\textbf{Params}} &
\textbf{F1$\uparrow$} &
\textbf{mIoU$\uparrow$} &
\makecell[c]{\textbf{OCR}\\\textbf{Acc.$\uparrow$}} &
\makecell[c]{\textbf{RO}\\\textbf{Score$\uparrow$}} \\
\midrule
Full FT & 1881.83M & 0.7336 & 0.8350 & 0.9408 & 0.9628 \\
Grid align only & 81.67M & 0.7541 & 0.8440 & 0.9566 & 0.9762 \\
Projector only & 35.96M & 0.6830 & 0.8139 & 0.9134 & 0.9262 \\
\textbf{Grid align + projector} & 83.99M & \textbf{0.9189} & \textbf{0.8919} & \textbf{0.9636} & \textbf{0.9774} \\
\bottomrule
\end{tabular}%
}
\end{minipage}
\hfill
\begin{minipage}[t]{0.47\textwidth}
\centering
\caption{Feature-layer selection ablation.}
\label{tab:layer_ablation}
\scriptsize
\setlength{\tabcolsep}{2.4pt}
\renewcommand{\arraystretch}{0.92}
\resizebox{\linewidth}{!}{%
\begin{tabular}{lcccc}
\toprule
\textbf{Layers} & \textbf{F1$\uparrow$} & \textbf{mIoU$\uparrow$} & \makecell[c]{\textbf{OCR}\\\textbf{Acc.$\uparrow$}} &
\makecell[c]{\textbf{RO}\\\textbf{Score$\uparrow$}} \\
\midrule
\{2, 5\} & 0.6855 & 0.7682 & 0.9076 & 0.9372 \\
\{16, 18, 20, 22\} & 0.7221 & 0.8175 & 0.9256 & 0.9412 \\
\textbf{\{2, 5, 8, 11\}} & \textbf{0.9192} & \textbf{0.8968} & \textbf{0.9602} & \textbf{0.9736} \\
\bottomrule
\end{tabular}%
}
\end{minipage}
\vspace{-2em}
\end{table}

Table~\ref{tab:spatial_ablation} isolates the contribution of each \adaptername{} component on Qwen3.5-2B. The full fine-tuning row is included as a reference, while all adapter rows keep the backbone frozen. Grid alignment alone provides a modest improvement over full fine-tuning, especially for RO and OCR. In contrast, projector-only training underperforms full fine-tuning, indicating that the gains do not come merely from adding trainable parameters. The full configuration, which combines grid-aware alignment with projector training, produces the largest improvement across all four structural metrics.

Table~\ref{tab:layer_ablation} studies which visual feature layers are most useful for the adapter. Early-to-mid layers, \{2,5,8,11\}, outperform both shallow-only layers and late-layer selections. This pattern is consistent with the design goal of preserving local spatial cues while retaining sufficient semantic abstraction for document elements. On both tested backbones, \adaptername{} improves localization relative to full fine-tuning, while its effects on reading order and description similarity vary. Overall, these ablations suggest that \adaptername{} primarily acts as a spatial-structure adapter rather than as a general-purpose language-generation module. This also explains why alt-text improvements are more mixed than the localization and reading-order gains.

\paragraph{Training-data scale.} On Qwen3.5-2B, \adaptername{} trails full fine-tuning on all four structural metrics at 10\% training data and yields mixed results at 30\%, while its structural advantage emerges at 50\% and 100\%. The complete comparison is reported in
Appendix~\ref{app:rebuttal_scaling}.

\subsection{Qualitative Results}
\label{subsec:qualitative_failure}
\begin{figure*}[t]
    \centering

    \begin{subfigure}[t]{0.49\textwidth}
        \centering
        \includegraphics[width=\linewidth]{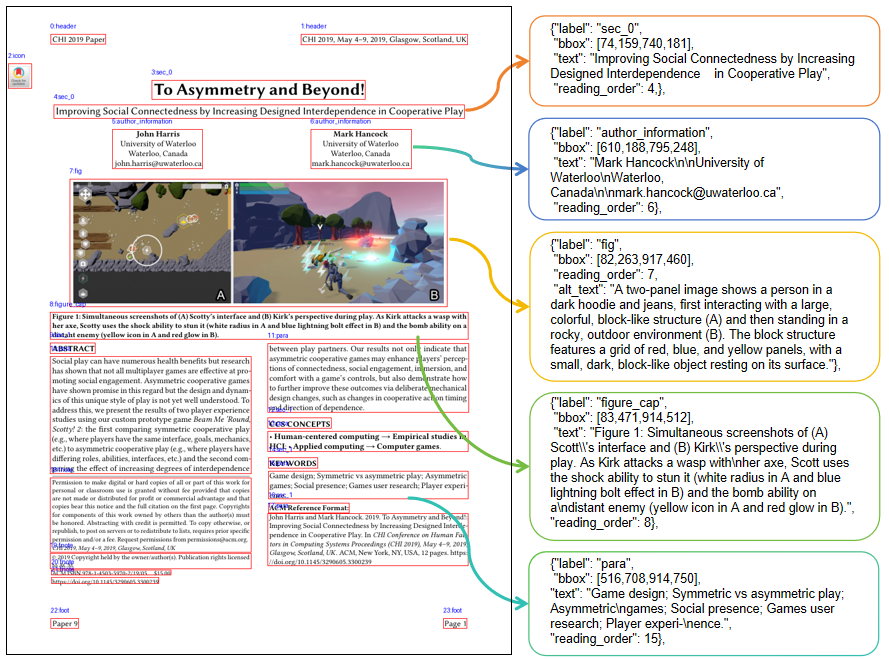}
        \caption{Ground truth}
        \label{fig:ground_truth}
    \end{subfigure}
    \hfill
    \begin{subfigure}[t]{0.49\textwidth}
        \centering
        \includegraphics[width=\linewidth]{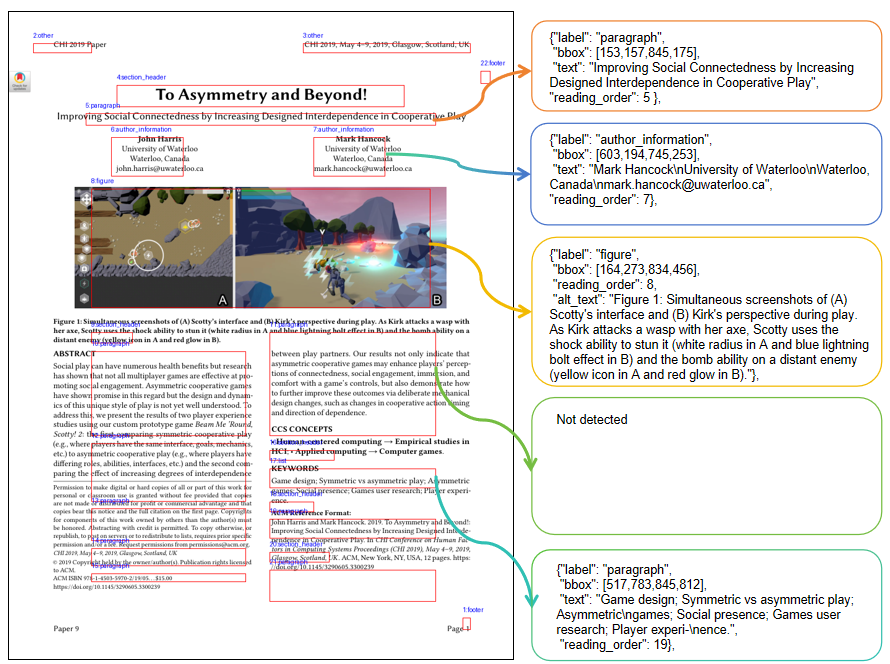}
        \caption{Qwen3.5-2B}
        \label{fig:qwen_base}
    \end{subfigure}

    \vspace{0.6em}

    \begin{subfigure}[t]{0.49\textwidth}
        \centering
        \includegraphics[width=\linewidth]{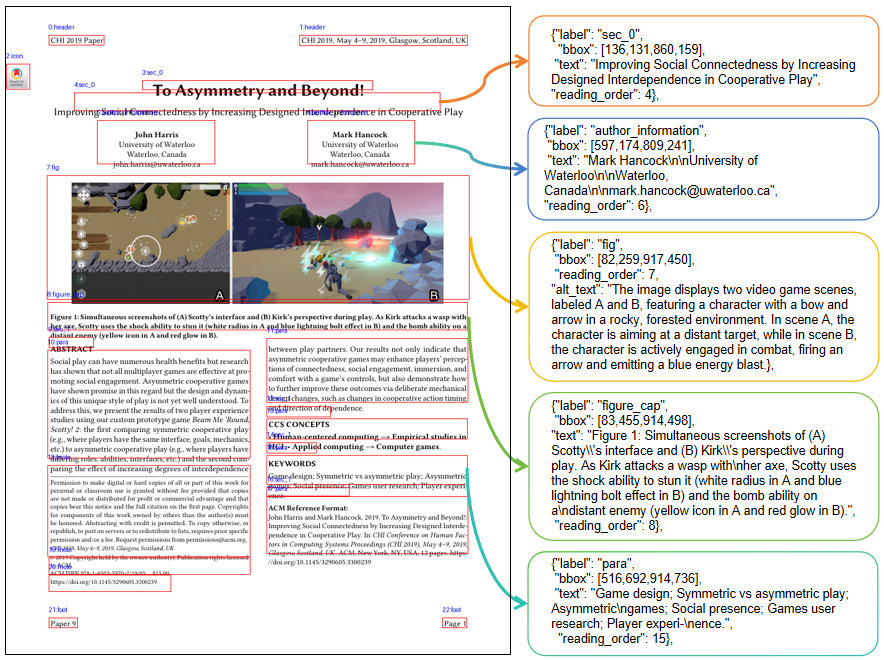}
        \caption{Qwen3.5-2B + full fine-tuning}
        \label{fig:qwen_ft}
    \end{subfigure}
    \hfill
    \begin{subfigure}[t]{0.49\textwidth}
        \centering
        \includegraphics[width=\linewidth]{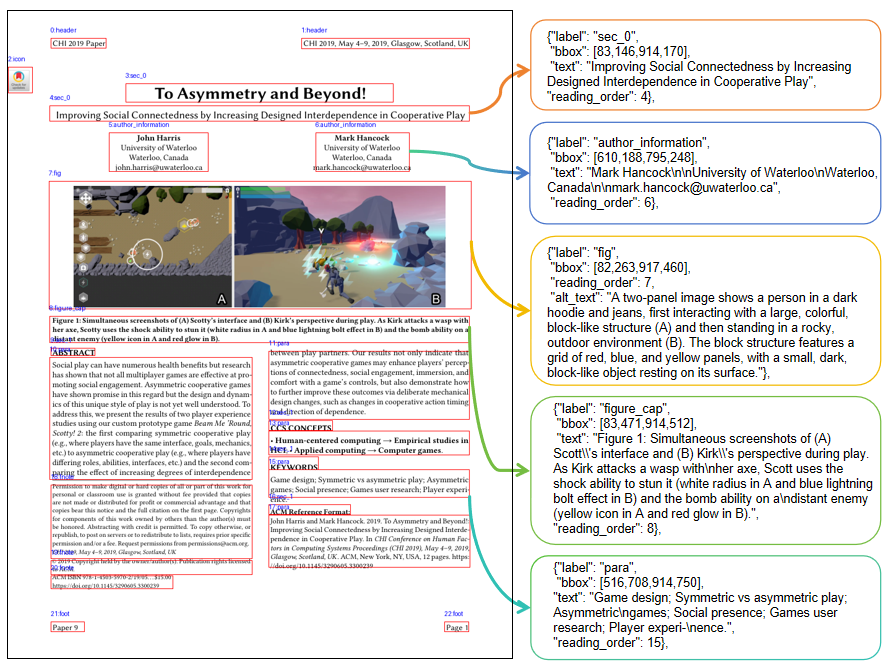}
        \caption{Qwen3.5-2B + \adaptername{}}
        \label{fig:qwen_adapter}
    \end{subfigure}

    \caption{
    Qualitative comparison of annotation mapping results. 
    All model-based methods use Qwen3.5-2B as the backbone. 
    We compare the ground-truth annotations with outputs from the base model, full fine-tuning, and the proposed \adaptername{} adapter.
    }
    \label{fig:output_examples}
\end{figure*}

Figure~\ref{fig:output_examples} compares annotation mapping results from the ground truth, the base Qwen3.5-2B model, full fine-tuning, and \adaptername{}. The base model often produces incomplete or poorly grounded mappings. Full fine-tuning improves schema compliance, but spatial alignment errors can remain. In comparison, \adaptername{} produces element mappings that are more consistent with the ground-truth page structure, especially for localized regions and their associated schema fields.

\section{Conclusion}
\label{sec:conclusion}

We introduced \datasetname{}, a unified benchmark for scientific-document parsing, and \adaptername{}, a spatial adapter tailored to the same page-level output contract. The task jointly evaluates localized elements, labels, reading order, textual and table content, caption links, and alternative text within a single ordered parse. Experiments show that existing systems tend to fail different parts in evaluation, while frozen-backbone \adaptername{} tuning substantially improves structural parsing for a general VLM. 

\noindent \textbf{Limitations.} \datasetname{} is limited to open-access ACM scientific PDFs and primarily covers the first eight pages, leaving late appendices and references underrepresented. Alt-text annotations rely on LLM assistance with human refinement, and closed-source pre-training exposure cannot be fully excluded. The benchmark is intended to support accessible document understanding, search, and document mining, but generated accessibility metadata should not be treated as deployment-ready without further validation.

\begin{ack}
This work was supported in part by Helmholtz Association of German Research Centers, in part by the Ministry of Science, Research and the Arts of Baden-Württemberg (MWK) through the Cooperative Graduate School Accessibility through AI-based Assistive Technology (KATE) under Grant BW6-03, and in part by the National Natural Science Foundation of China under Grant No. 62503166, in part by the Hunan Provincial Research and Development Project under Grant number 2026QK3018, and in part by the Yuelushan Industrial Innovation Center. This work was partially performed on the HoreKa supercomputer funded by the MWK and by the Federal Ministry of Education and Research, partially on the HAICORE@KIT partition supported by the Helmholtz Association Initiative and Networking Fund, and partially on bwForCluster Helix supported by the state of Baden-Württemberg through bwHPC and the German Research Foundation (DFG) through grant INST 35/1597-1 FUGG. 
\end{ack}


{
\small
\bibliographystyle{unsrt}
\bibliography{main}
}

\clearpage

\appendix

\section{Dataset Details}
\label{app:dataset_details}

This appendix is supplementary reading. The main text defines the task contract and reports the primary experimental conclusions; the material below gives the distributional and procedural detail needed to inspect the benchmark more closely.

\subsection{Label Taxonomy and Corpus Profile}

\datasetname{} uses a 23-label taxonomy rather than a small generic layout set. The goal is to keep elements that matter for reading order, reflow, caption linkage, and auditability visible as distinct targets. Headings, paragraph fragments, captions, equations, code blocks, metadata, references, lists, and visual objects therefore remain separately labeled.

\begin{table}[htbp]
\centering
\caption{Semantic label taxonomy used in Read-Parse-Describe.}
\label{tab:semantic_labels_double}
\small 
\begin{tabular}{@{}cll|cll@{}}
\midrule
\textbf{ID} & \textbf{Description} & \textbf{Label} & \textbf{ID} & \textbf{Description} & \textbf{Label} \\ \midrule
1  & Paper Title        & \texttt{sec\_0}      & 13 & Table              & \texttt{tab}         \\
2  & Level-1 Heading    & \texttt{sec\_1}      & 14 & Figure-Caption     & \texttt{figure-cap}  \\
3  & Level-2 Heading    & \texttt{sec\_2}      & 15 & Table-Caption      & \texttt{table-cap}   \\
4  & Level-3 Heading    & \texttt{sec\_3}      & 16 & Annotation         & \texttt{anno}        \\
5  & Level-4 Heading    & \texttt{sec\_4}      & 17 & Formula            & \texttt{equ}         \\
6  & Paragraph          & \texttt{para}       & 18 & Code Block         & \texttt{code}        \\
7  & Spanning Paragraph & \texttt{half\_para} & 19 & Catalog            & \texttt{catalogue}   \\
8  & Header             & \texttt{header}     & 20 & Reference          & \texttt{reference}   \\
9  & Footer             & \texttt{foot}       & 21 & Author-information & \texttt{author-info} \\
10 & Footnote           & \texttt{fnote}      & 22 & Conference logo    & \texttt{icon}        \\
11 & Watermark          & \texttt{watermark}  & 23 & List               & \texttt{list}        \\
12 & Figure             & \texttt{fig}        &    &                    &                      \\ \bottomrule
\end{tabular}
\end{table}

The benchmark itself contains 2,402 curated scientific documents and 18,163 continuous pages. The domain chart below summarizes the broader analyzed source-corpus pool of 2,151 ACM articles from 2020--2024; the residual slice groups domains that were not individually named in the figure.

\begin{table}[ht]
\centering
\caption{Primary Domain Distribution of Analyzed Articles from 2020 - 2024}
\label{tab:analysis Articles}
\begin{tabular}{l|r|r}
\hline
\textbf{Category} & \textbf{Count} & \textbf{Percentage} \\ \midrule
AI \& ML & 271 & 12.60\% \\ \hline
Accessibility & 255 & 11.85\% \\ \hline
AR/VR/XR & 235 & 10.93\% \\ \hline
Design \& HCI Methods & 149 & 6.93\% \\ \hline
Privacy \& Social & 103 & 4.79\% \\ \hline
Health \& Wellbeing & 95 & 4.42\% \\ \hline
Demographics & 71 & 3.30\% \\ \hline
Games, EdTech \& Creativity & 65 & 3.02\% \\ \hline
Ethics, Fairness \& CSCW & 61 & 2.84\% \\ \hline
Hardware \& Fabrication & 56 & 2.60\% \\ \hline
Others / Long-tail & 790 & 36.72\% \\ \midrule \hline
\textbf{Total} & \textbf{2,151} & \textbf{100.00\%} \\ \bottomrule \hline
\end{tabular}
\end{table}

\subsection{Page Curation and Element Distributions}

The first-eight-page rule is intended to increase the density of multimodal scientific content while keeping each paper as a continuous page sequence. It preserves the early-paper regions where abstracts, methods, figures, tables, captions, equations, and author metadata interact most strongly. Later pages are not treated as unimportant; they are underrepresented because they are disproportionately reference-heavy in the analyzed full-text subset.

\begin{table}[htbp]
\centering
\small
\caption{Impact of the 8-Page Truncation Strategy on Layout Element Distribution. By comparing the element proportions in Pages 1--8 versus Pages $\ge 9$ (estimated on a full-text subset), we observe that the truncation effectively preserves high-value multimodal objects while filtering out visually monotonous reference lists.}
\label{tab:truncation_bias}
\begin{tabular}{@{}lcccc@{}}
\toprule
\textbf{Element Category} & \textbf{Pages 1--8 (\%)} & \textbf{Pages $\ge 9$ (\%)} & \textbf{Density (1--8)} & \textbf{Density ($\ge 9$)} \\ \midrule
Title, Abstract \& Metadata & 100.00\% & 0.00\% & 1.2 / page & 0.0 / page \\
Equations \& Math Blocks & 78.42\% & 21.58\% & 4.5 / page & 1.2 / page \\
Figures \& Captions & 69.15\% & 30.85\% & 2.1 / page & 0.9 / page \\
Body Paragraphs & 62.30\% & 37.70\% & 12.4 / page & 7.5 / page \\
Tables \& Algorithms & 54.80\% & 45.20\% & 0.8 / page & 0.7 / page \\
References & 8.25\% & 91.75\% & 0.5 / page & 15.6 / page \\ \midrule
\textbf{Overall Multimodal Complexity} & \textbf{High} & \textbf{Low} & \textbf{Diverse} & \textbf{Monotonous} \\ \bottomrule
\end{tabular}
\end{table}

The following tables decompose the visual and layout population from two complementary angles. Subject categories describe the broad disciplinary context of visual instances. Image types describe the visual form that the alt-text and caption-linking steps must handle. Aggregated layout groups show how much of the annotation budget is spent on document structure, metadata, scholarly objects, captions, references, and fragments.

\begin{table}[ht]
\centering
\caption{Topic Type Distribution}
\label{tab:categories}
\begin{tabular}{l|r|r}
\hline
\textbf{Category} & \textbf{Count} & \textbf{Percentage} \\ \midrule
Computer Science \& AI & 6,443 & 56.8\% \\  \hline
Social Science & 2,181 & 19.2\% \\ \hline
Engineering & 1,359 & 12.0\% \\ \hline
Medicine \& Healthcare & 652 & 5.8\% \\ \hline
Other & 245 & 2.2\% \\ \hline
Biology & 110 & 1.0\% \\ \hline
Physics & 106 & 0.9\% \\ \hline
Mathematics & 98 & 0.9\% \\ \hline
Economics \& Finance & 72 & 0.6\% \\ \hline
Environmental Science & 50 & 0.4\% \\ \hline
Chemistry & 22 & 0.2\% \\ \midrule \hline
\textbf{Total} & \textbf{11,338} & \textbf{100.0\%} \\ \bottomrule \hline
\end{tabular}
\end{table}
\begin{table}[ht]
\centering
\caption{Image Type Distribution}
\label{tab：image_type}
\begin{tabular}{l|r|r}
\toprule
\textbf{Image Type} & \textbf{Count} & \textbf{Percentage} \\ \midrule
Table & 2758 & 24.3\% \\ \hline
Diagrammatic Instruction / Concept & 2086 & 18.4\% \\ \hline
User Interface / Screenshot & 1743 & 15.4\% \\ \hline
Photographic Image & 1529 & 13.5\% \\ \hline
Flowchart / Block Diagram & 1000 & 8.8\% \\ \hline
Bar Chart / Histogram & 942 & 8.3\% \\ \hline
Line Graph & 466 & 4.1\% \\ \hline
Other & 371 & 3.3\% \\ \hline
Scatter Plot & 205 & 1.8\% \\ \hline
Heatmap & 147 & 1.3\% \\ \hline
Pie Chart & 27 & 0.2\% \\ \hline
Anatomical Illustration & 27 & 0.2\% \\ \hline
Circuit Diagram & 21 & 0.2\% \\ \hline
Radar Chart & 12 & 0.1\% \\ \hline
Molecular Structure & 4 & 0.0\% \\ \hline \hline
\textbf{Total} & \textbf{11338} & \textbf{100.0\%} \\ \hline
\end{tabular}
\end{table}

\begin{table}[htbp]
\centering
\caption{Distribution of high-value layout elements.}
\label{tab:label_dist}
\begin{tabular}{@{}lccc@{}}
\toprule
\textbf{Category} & \textbf{Label} & \textbf{Count} & \textbf{Percentage} \\ \midrule
Structural Headers & \texttt{sec\_1}, \texttt{sec\_2}, \texttt{sec\_0} & 44,617 & 35.64\% \\
Academic Metadata & \texttt{author-info}, \texttt{fnote} & 28,381 & 22.66\% \\
Scholarly Objects & \texttt{fig}, \texttt{tab}, \texttt{equ}, \texttt{code} & 12,336 & 9.85\% \\
Captions and References & \texttt{figure-cap}, \texttt{table-cap}, \texttt{reference} & 21,381 & 17.08\% \\
Layout Fragments & \texttt{half\_para}, \texttt{list}, \texttt{icon} & 17,564 & 14.03\% \\
Others & \texttt{anno}, \texttt{sec\_3}, \texttt{sec\_4}, \texttt{watermark}, etc. & 939 & 0.74\% \\
\midrule
\textbf{Total} & -- & \textbf{125,218} & \textbf{100.00\%} \\
\bottomrule
\end{tabular}
\end{table}

\section{Alt-text Annotation Details}
\label{app:alttext}

Visual descriptions are harder to audit than boxes or OCR strings because a fluent sentence can still be vague, hallucinated, or attached to the wrong object. We therefore separate the alt-text process into generation, rubric-based scoring, targeted human refinement, and a manual audit. The detailed prompt and rubric below are provided to make the label assumptions inspectable.

\begin{table*}[htbp]
\centering
\small
\caption{Detailed Prompt Configurations and Expanded Evaluation Rubric for Alt-text Generation}
\label{tab:alttext_generation_eval_expanded}
\renewcommand{\arraystretch}{1.4}
\begin{tabularx}{\textwidth}{>{\raggedright\arraybackslash}p{3.5cm} >{\raggedright\arraybackslash}X}
\toprule
\textbf{Configuration} & \textbf{Detailed Content / Implementation} \\
\midrule
\textbf{Generation Prompt} 
& ``Describe this image in 1--3 concise sentences for accessibility. If the image contains multiple panels or subfigures, summarize them as a single combined figure rather than describing each panel separately. Focus only on the key visual information, such as the type of chart or image, the main subjects, and clearly visible labels or visual patterns. Do not use phrases like 'image of' or 'picture of'. Provide a factual description based only on what is visible, without interpreting or explaining the underlying data.'' \\
\midrule
\textbf{Generation Parameters} 
& \textbf{Temperature:} 0.2 \quad \textbf{Top-p:} 0.9 \quad \textbf{Max New Tokens:} 256 \quad \textbf{Repetition Penalty:} 1.05 \\
\midrule
\textbf{Evaluation Role} 
& \textbf{System Prompt:} You are a rigorous Image Accessibility Auditor specializing in scientific and technical content. Your goal is to determine if the provided 'Alt-text' accurately represents the visual information for a visually impaired user. Evaluate strictly based on Accuracy, Completeness, and Constraint Adherence. \\
\midrule
\textbf{Evaluation Rubric} 
& \textbf{Score 1 (Failure - Misleading/Hallucination):} Completely irrelevant, contains gross data hallucinations, or generates pure technical noise. Fails to identify the basic type of the image. Actively misleads the visually impaired user. \newline
\newline
\textbf{Score 2 (Poor - Severe Omissions/Violations):} Identifies the general chart/image type but suffers from severe omissions (e.g., missing critical axes, legends, or main subjects). Alternatively, it severely violates negative constraints (e.g., provides subjective interpretations of data instead of visual facts). \newline
\newline
\textbf{Score 3 (Acceptable - Basic but Flawed):} Correctly identifies the main subjects and chart type. However, it lacks specific critical details (e.g., distinct visual trends or key labels). May fail minor constraints (e.g., exceeds the 3-sentence limit, or includes filler phrases like "This is a picture of"). \newline
\newline
\textbf{Score 4 (Good - Professional \& Factual):} Clear, objective, and accurate. Successfully captures the chart type, major variables, key labels, and significant visual patterns. Strictly adheres to negative constraints. May only miss minor, non-essential visual nuances or slightly lack extreme conciseness. \newline
\newline
\textbf{Score 5 (Excellent - Perfect Accessibility):} Highly concise (strictly 1-3 sentences), factual, and deeply informative. Seamlessly summarizes complex or multi-panel figures into a cohesive whole. Provides a visually impaired user with the exact informational equivalent of a quick visual scan, devoid of any over-interpretation or fluff. \\
\bottomrule
\end{tabularx}
\end{table*}

The score distribution reports the same 11,338 visual instances before and after human refinement. Low-scoring generated descriptions are revised, then re-scored under the same rubric.

\begin{table}[ht]
\centering
\caption{Alt text generation: LLM-generated vs LLM+human\_modified}
\begin{tabular}{l|c|c}
\hline
\textbf{category} & \textbf{LLM generated (Count)} & \textbf{LLM + human\_modified (Count)} \\ \hline
Score 1 & 0 & 0 \\ \hline
Score 2 & 146 & 0 \\ \hline
Score 3 & 1262 & 0 \\ \hline
Score 4 & 2419 & 226 \\ \hline
Score 5 & 7511 & 11112 \\ \hline \hline
\textbf{Score 5} & \textbf{66.2\%} & \textbf{98.1\%} \\ \hline
\end{tabular}
\end{table}

\begin{figure}[t]
    \centering
    \includegraphics[width=0.86\textwidth]{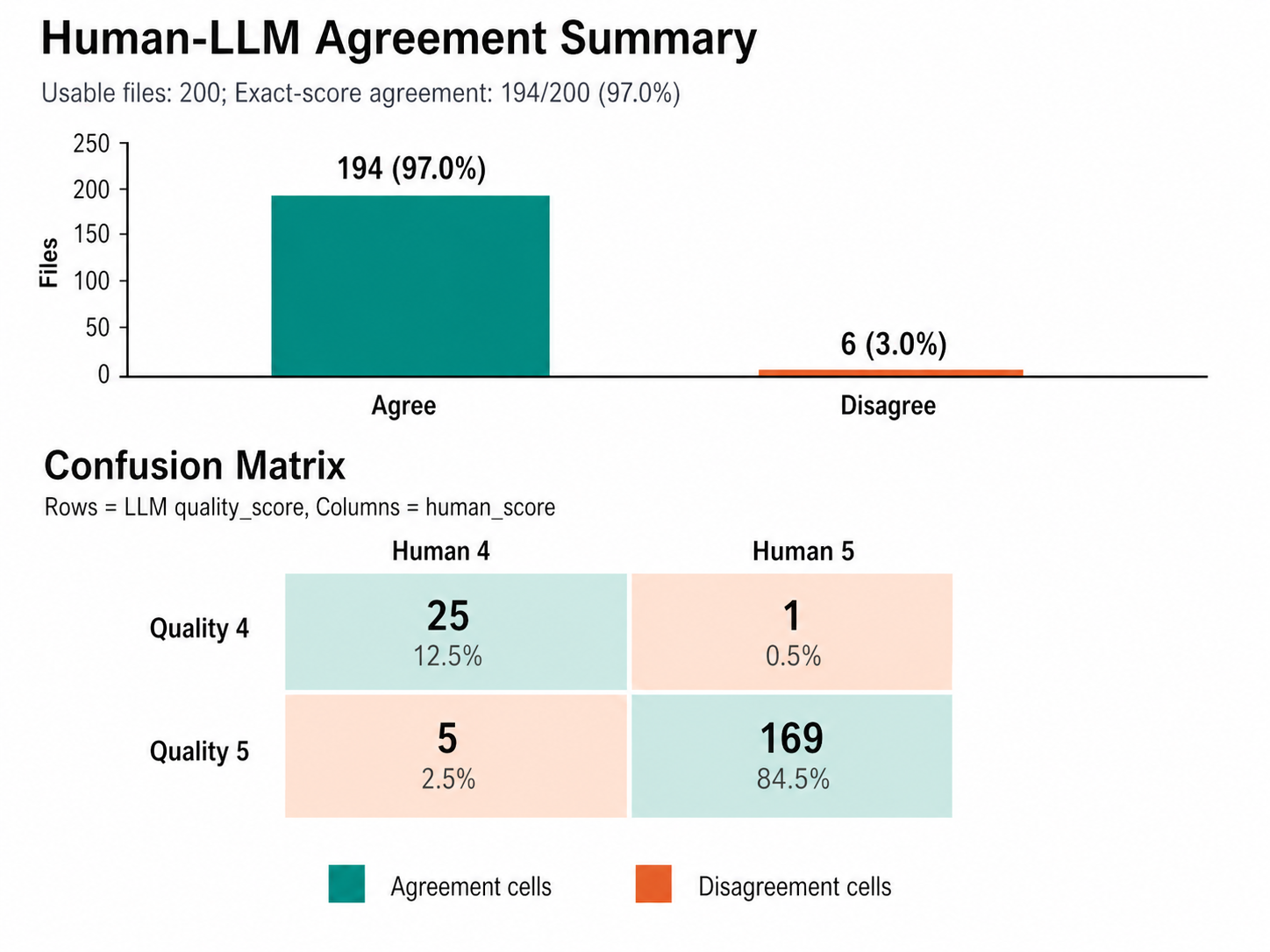}
    \caption{Human-LLM agreement audit over 200 revised samples. Agreement is measured after low-score descriptions are refined and re-scored.}
    \label{fig:agreement_audit}
\end{figure}

\section{Evaluation Metric}
To quantitatively assess the performance of the fine-tuned models in handling complex scholarly layouts, we employ a multi-dimensional evaluation suite. This suite covers three complementary aspects: (1) \textit{physical localization} of layout elements, (2) \textit{logical sequencing} of the reading flow, and (3) \textit{textual and tabular accuracy} of the extracted content.(4) \textit{Accessibility Metadata Evaluation}.Such a decomposition is necessary because a model can, for example, detect elements correctly but still fail to order them properly, or produce fluent text while misplacing bounding boxes.

\vspace{\baselineskip} 
\noindent\textbf{1. Physical Layout Analysis: F1-score and mIoU}
\vspace{\baselineskip} 

For Document Layout Analysis (DLA), we evaluate the model's ability to correctly identify and localize structural entities (e.g., figures, tables, and section blocks).

\paragraph{F1-score.}
We compute the harmonic mean of Precision and Recall for each label category. A detection is considered a True Positive (TP) if (i) the predicted category matches the ground-truth category and (ii) the overlap between predicted and ground-truth regions exceeds a predefined threshold (typically based on IoU). The metric is defined as:
$$F1 = 2 \cdot \frac{Precision \cdot Recall}{Precision + Recall}$$
In practice, F1-score summarizes the trade-off between missing elements (low Recall) and producing spurious detections (low Precision).

\paragraph{mIoU (mean Intersection over Union).}
To measure geometric localization accuracy, we compute the Intersection over Union (IoU) between each predicted bounding box $B_p$ and the corresponding ground-truth box $B_{gt}$. The mean IoU over $N$ matched instances is reported as:
$$mIoU = \frac{1}{N} \sum_{i=1}^{N} \frac{Area(B_{p,i} \cap B_{gt,i})}{Area(B_{p,i} \cup B_{gt,i})}$$
Compared to F1-score, mIoU is more sensitive to precise boundary alignment, which is critical for downstream tasks such as figure-to-caption association or table cell reconstruction.

\vspace{\baselineskip} 
\noindent\textbf{2. Logical Reading Order: Normalized Edit Distance (NED)}
\vspace{\baselineskip} 

To evaluate Reading Order (RO), we represent the sequence of detected blocks as a string of identifiers and compare it with the ground-truth sequence. We use the Levenshtein edit distance to measure the minimal number of insertions, deletions, and substitutions needed to transform the predicted order into the correct order. To ensure comparability across documents with different numbers of blocks, we normalize the distance and define the RO score as:
$$RO_{Score} = 1 - \frac{Levenshtein(S_p, S_{gt})}{\max(|S_p|, |S_{gt}|)}$$
where $S_p$ and $S_{gt}$ denote the predicted and ground-truth block sequences, respectively. A higher score indicates a more faithful logical flow, which is essential for reflowable rendering and accessible reading.

\vspace{\baselineskip} 
\noindent\textbf{3. Textual and Tabular Accuracy: ED and TEDS}
\vspace{\baselineskip} 

\paragraph{OCR accuracy (Edit Distance, ED).}
For general text extraction, we compute character-level edit distance between the recognized text and the reference transcription. This metric directly reflects transcription fidelity and is particularly informative for scientific documents containing symbols, subscripts, and uncommon terminology.

\paragraph{Table reconstruction (TEDS).}
For table parsing, we use the Tree Edit Distance Based Similarity (TEDS), which evaluates both the global structure (row/column organization) and the cell content by representing tables as tree structures. The similarity is defined as:
$$TEDS(T_a, T_b) = 1 - \frac{Dist_{tree}(T_a, T_b)}{\max(|T_a|, |T_b|)}$$
where $T_a$ and $T_b$ are the predicted and ground-truth table trees. This metric is more aligned with human judgment than plain text metrics, because it penalizes structural errors (e.g., merged/split cells) in addition to textual mistakes.

\vspace{\baselineskip} 
\noindent\textbf{4. Accessibility Metadata Evaluation: METEOR Score and BERTScore}
\vspace{\baselineskip} 

To evaluate the quality of the generated Alt-text for figures and tables, we employ the METEOR  and BERTScore metrics. While metrics like BLEU primarily focus on exact n-gram overlaps, METEOR and BERTScore are designed to capture the semantic nuances of scholarly descriptions more effectively.
\paragraph{The Rationale for Choosing METEOR}
In the context of academic document reconstruction, the description of a figure (Alt-text) can be linguistically diverse. METEOR is superior for this task due to its unique features: Synonym Matching: It utilizes a dictionary (like WordNet) to match synonyms. For instance, if the ground truth says "Diagram" and the model generates "Schematic," METEOR recognizes this as a valid match, whereas BLEU would penalize it. Stemming: It matches words with the same root (e.g., "clustering" and "clustered"), which is crucial for maintaining the technical accuracy of alt-text.Alignment-based Scoring: It creates an explicit alignment between the predicted string and the reference, penalizing incorrect word orders through a "fragmentation penalty."
\paragraph{Mathematical Formulation}
The METEOR score is calculated by combining the harmonic mean of precision ($P$) and recall ($R$), with a specific emphasis on recall, followed by a penalty for incorrect ordering. First, the F-score is computed as:$$F_{mean} = \frac{P \cdot R}{\alpha \cdot P + (1 - \alpha) \cdot R}$$Typically, $\alpha$ is set to give more weight to recall, ensuring that the descriptive content of the figure is fully captured. Then a Penalty ($Pen$) is calculated based on the "chunkiness" of the alignment (how fragmented the matches are):$$Pen = \gamma \cdot \left( \frac{\#chunks}{\#unigrams\_matched} \right)^\beta$$The final METEOR Score is:$$Score = F_{mean} \cdot (1 - Pen)$$
\paragraph{The Rationale for Choosing BERTScore}
While METEOR effectively accounts for lexical and morphological variations, scholarly descriptions often convey identical visual evidence through substantially different syntactic patterns. BERTScore is selected to capture this deep semantic equivalence beyond surface-level n-gram overlap. By generating contextualized token embeddings from pre-trained language models, it dynamically evaluates polysemous technical terms within their precise scholarly contexts. Furthermore, BERTScore relies on continuous pairwise cosine similarities rather than rigid discrete lookups, which fairly credits semantically rich paraphrases without external dictionaries. This focus on contextualized conceptual alignment rather than rigid syntactic templates ensures a reliable assessment of whether the essential semantic takeaways of complex scientific figures are preserved for visually impaired readers.

\paragraph{Mathematical Formulation}
BERTScore computes greedy token matching over contextualized representations. Let $\hat{x} = \langle \hat{x}_1, \dots, \hat{x}_l \rangle$ and $x = \langle x_1, \dots, x_k \rangle$ denote the generated description and ground-truth reference, with $\mathbf{\hat{x}}_j$ and $\mathbf{x}_i$ representing their respective pre-trained normalized token embeddings. The pairwise cosine similarities are greedily aligned to derive contextual recall ($R_{BERT}$) and precision ($P_{BERT}$):
\begin{equation}
R_{BERT} = \frac{1}{|x|} \sum_{x_i \in x} \max_{\hat{x}_j \in \hat{x}} \mathbf{x}_i^\top \mathbf{\hat{x}}_j, \quad P_{BERT} = \frac{1}{|\hat{x}|} \sum_{\hat{x}_j \in \hat{x}} \max_{x_i \in x} \mathbf{x}_i^\top \mathbf{\hat{x}}_j.
\end{equation}
The final reported semantic metric is the harmonic mean ($F_{BERT}$) combining both directional alignments:
\begin{equation}
F_{BERT} = 2 \cdot \frac{P_{BERT} \cdot R_{BERT}}{P_{BERT} + R_{BERT}}.
\end{equation}

\section{Evaluation Details}
\label{app:evaluation_details}

All reported systems are converted into the same element schema before scoring. Layout matching requires category agreement and an IoU threshold of 0.5. DLA-F1 is computed over matched layout elements, and mIoU averages the IoU of true-positive matches. Reading order is evaluated over matched element identifiers, so a model must both find the relevant element and place it correctly in the sequence. OCR is normalized before edit-distance scoring, and table content is scored through the serialized \texttt{html} field when a system produces it. Visual descriptions are scored only for visual instances with reference alternative text; a missing description for an applicable object receives zero for the description metric.

Unparsable predictions are not discarded globally. Minor syntax errors are repaired deterministically when the intended field structure is recoverable. If a prediction remains unparsable, only the affected fields are treated as empty. This localizes a syntax failure to the fields it prevents from being evaluated and avoids giving a model credit for content that cannot be reliably associated with an element object.
\begin{table}[htbp]
\centering
\small
\caption{Hyperparameters and implementation details for model fine-tuning.}
\label{tab:training_details}
\begin{tabular}{ll}
\toprule
\textbf{Hyperparameter / Setting} & \textbf{Value} \\
\midrule
\multicolumn{2}{l}{\textit{Model \& Strategy}} \\
\midrule
Base Model & Qwen3.5-2B \\
Training Strategy & Full Fine-Tuning \\
Trainable Modules & Aligner, LLM \\
Frozen Modules & ViT (\texttt{freeze\_vit=True}) \\
Precision & bfloat16 \\
Max Sequence Length & 8192 \\
\midrule
\multicolumn{2}{l}{\textit{Optimization}} \\
\midrule
Training Epochs & 1 \\
Peak Learning Rate & 2e-5 \\
Optimizer & AdamW ($\beta_1=0.9, \beta_2=0.999$) \\
LR Scheduler & Cosine Decay \\
Batch Size per Device & 2 \\
Gradient Accumulation Steps & 4 \\
Effective Batch Size & $8 \times \text{Number of GPUs}$ \\
\midrule
\multicolumn{2}{l}{\textit{Data \& Environment}} \\
\midrule
Data Augmentation & Dynamic Cropping \\
Hardware & $1 \times \text{NVIDIA A100 (40GB)}$ \\
Peak VRAM per GPU & ${\sim}38.5$~GB \\
Total Training Time & ${\sim}20$--$24$~Hours \\
\bottomrule
\end{tabular}
\end{table}
To ensure reproducibility and provide a clear overview of our experimental setup, the primary training hyperparameters, optimization configurations, and computing resources for model fine-tuning are summarized in Table~\ref{tab:training_details}. Unless specified otherwise, other baseline models and our proposed adapter follow comparable training schedules and hardware environments for fair comparison.
\section{Training-data Scaling}
\label{app:rebuttal_scaling}

We compare full fine-tuning and \adaptername{} on Qwen3.5-2B using
10\%, 30\%, 50\%, and 100\% of the training data under the same reported
evaluation protocol. Table~\ref{tab:rebuttal_scaling} reports all six metrics.
At 10\%, \adaptername{} trails full fine-tuning on all four structural
metrics. Results are mixed at 30\%; at 50\%, it improves F1, mIoU, and OCR
while matching RO, and at 100\% it improves all four.
The F1 difference changes from $-0.0756$ at 10\% to $+0.1790$ at 50\%
and $+0.1853$ at 100\%. Thus, the structural advantage emerges at larger
training fractions in this experiment. METEOR favors the adapter at
each fraction, while BERTScore changes are mixed.


\begin{table}[t]
\centering
\scriptsize
\setlength{\tabcolsep}{4pt}
\caption{Training-data scaling on Qwen3.5-2B. Fractions indicate the
reported proportion of the training data.}
\label{tab:rebuttal_scaling}
\begin{tabular}{llrrrrrr}
\toprule
Data & Method & F1 & mIoU & OCR & RO & METEOR & BERTScore \\
\midrule
10\% & Full FT & 0.6487 & 0.7563 & 0.8476 & 0.7996 & 0.1256 & 0.8106 \\
10\% & SARA & 0.5731 & 0.7144 & 0.7311 & 0.7805 & 0.1672 & 0.8002 \\
30\% & Full FT & 0.6771 & 0.7861 & 0.8756 & 0.8504 & 0.2651 & 0.8288 \\
30\% & SARA & 0.6567 & 0.8061 & 0.8087 & 0.8226 & 0.2871 & 0.8372 \\
50\% & Full FT & 0.6991 & 0.8232 & 0.9362 & 0.9623 & 0.3072 & 0.8540 \\
50\% & SARA & 0.8781 & 0.8533 & 0.9562 & 0.9623 & 0.3340 & 0.8533 \\
100\% & Full FT & 0.7336 & 0.8350 & 0.9408 & 0.9628 & 0.3741 & 0.8928 \\
100\% & SARA & 0.9189 & 0.8919 & 0.9636 & 0.9774 & 0.3834 & 0.8946 \\
\bottomrule
\end{tabular}
\end{table}

\section{Resolution Sensitivity}
\label{app:rebuttal_resolution}

We evaluate Qwen3.5-2B+\adaptername{} at 100\%, 50\%, and 25\% of the
reported native crop resolution while keeping the model and evaluation
settings fixed. As shown in Table~\ref{tab:rebuttal_resolution}, reducing
resolution to 25\% decreases F1 by 0.1600, mIoU by 0.0725, OCR by 0.1988,
and RO by 0.0910. Because mIoU is averaged over matched true positives,
its remaining value describes overlap among surviving matches and does
not indicate preservation of all element boundaries. This experiment
measures absolute input-resolution sensitivity; it does not establish
a robustness advantage over full fine-tuning without a corresponding
control evaluated at the same resolutions.

\begin{table}[t]
\centering
\small
\caption{Resolution sensitivity of Qwen3.5-2B+SARA under the
reported crop-resolution.}
\label{tab:rebuttal_resolution}
\begin{tabular}{lrrrr}
\toprule
Resolution & F1 & mIoU & OCR & RO \\
\midrule
100\% & 0.9189 & 0.8919 & 0.9636 & 0.9774 \\
50\% & 0.8389 & 0.8673 & 0.9248 & 0.9534 \\
25\% & 0.7589 & 0.8194 & 0.7648 & 0.8864 \\
\bottomrule
\end{tabular}
\end{table}

\section{Failure cases.}
\begin{figure*}[t]
    \centering

    \begin{subfigure}[b]{0.48\textwidth}
        \centering
        \includegraphics[width=\linewidth]{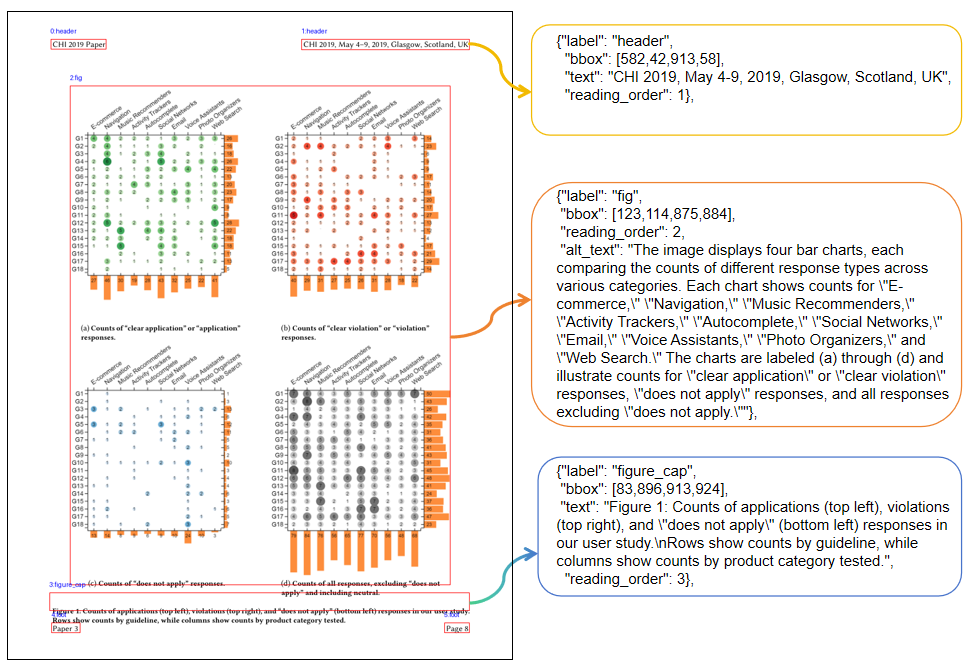}
        \caption{Prediction: borderless composite figure}
        \label{fig:failure_case1_pred}
    \end{subfigure}
    \hfill 
    \begin{subfigure}[b]{0.48\textwidth}
        \centering
        \includegraphics[width=\linewidth]{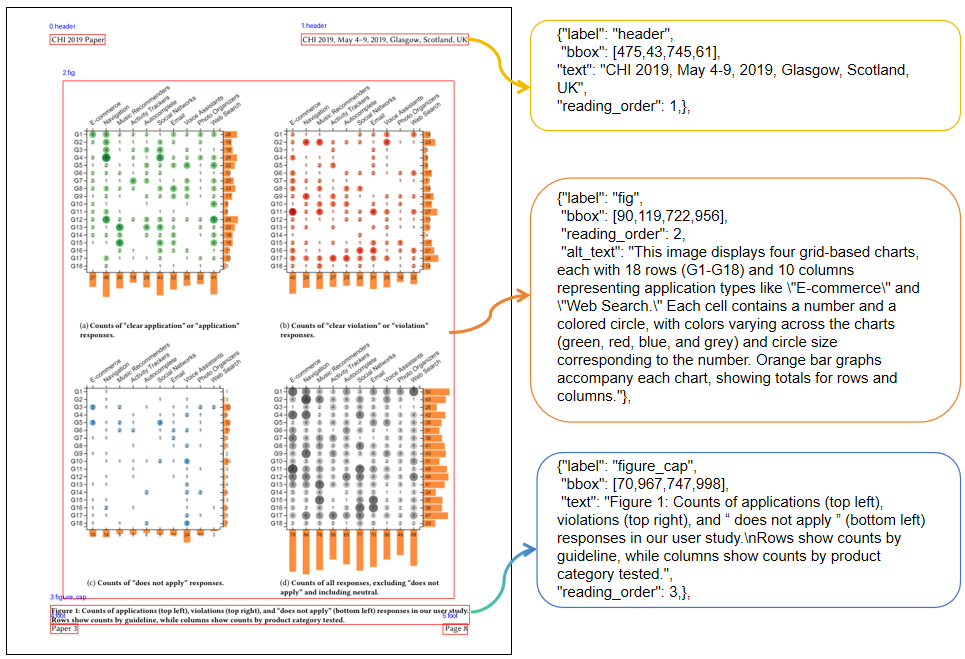} 
        \caption{Ground truth for Case 1}
        \label{fig:failure_case1_gt}
    \end{subfigure}

    \vspace{0.6em}

    \begin{subfigure}[b]{0.48\textwidth}
        \centering
        \includegraphics[width=\linewidth]{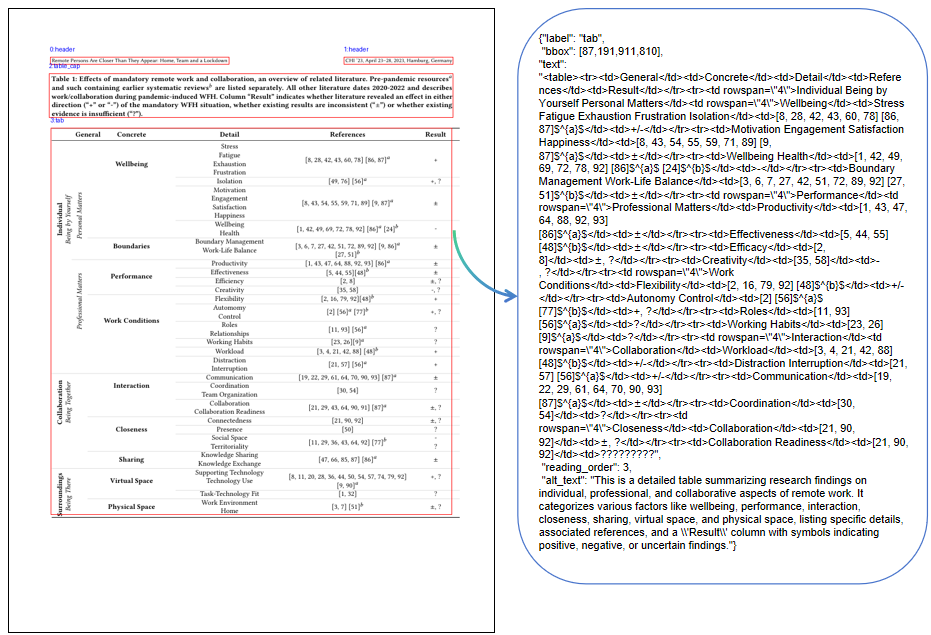}
        \caption{Prediction: dense table}
        \label{fig:failure_case2_pred}
    \end{subfigure}
    \hfill 
    \begin{subfigure}[b]{0.48\textwidth}
        \centering
        \includegraphics[width=\linewidth]{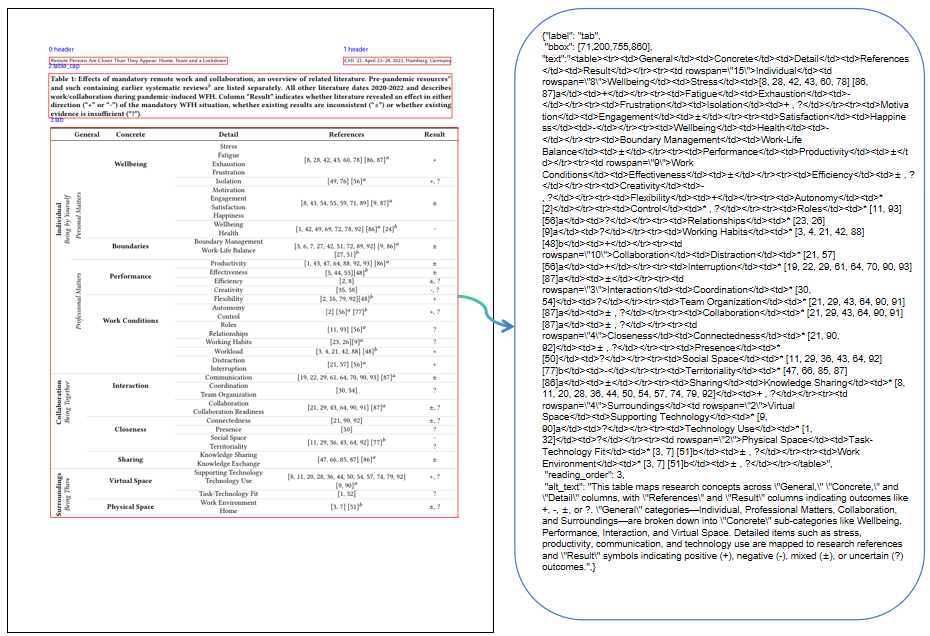} 
        \caption{Ground truth for Case 2}
        \label{fig:failure_case2_gt}
    \end{subfigure}
    
    \caption{
    Representative failure cases of \adaptername{}.
    Top row: a borderless composite figure where intra-image text can cause a spatial shift between figure regions and their captions.
    Bottom row: a dense table where localization remains relatively accurate, but decoding may truncate long content and introduce minor hallucinated artifacts.
    }
    \label{fig:failure_cases}
\end{figure*}

To better understand the limitations of \adaptername{}, we analyze two representative failure modes, shown in Figure~\ref{fig:failure_cases}. First, borderless composite figures can confuse the model when multiple sub-figures and intra-image text are densely arranged. In such cases, visual and textual cues become entangled, which may shift predicted figure boxes away from their corresponding captions. Second, excessively dense tables can stress the decoding process. Although the table region may still be localized accurately, long serialized outputs may be truncated near the end, and the model may occasionally introduce minor irrelevant artifacts.

\end{document}